\pdfoutput=1

\documentclass[11pt]{article}

\usepackage[final]{EMNLP2023}

\usepackage{times}
\usepackage{latexsym}

\usepackage[T1]{fontenc}

\usepackage[utf8]{inputenc}

\usepackage{microtype}

\usepackage{inconsolata}

\usepackage{babel}
\usepackage{soul}
\usepackage{xspace}
\usepackage{stmaryrd}
\usepackage{booktabs}
\usepackage{url}
\usepackage{color}
\usepackage{makecell}
\usepackage{bbm}
\usepackage{amsmath}
\usepackage{amssymb}
\usepackage{amsfonts} 
\usepackage{mathtools}
\usepackage{arydshln}
\usepackage{subcaption}
\usepackage{caption}
\usepackage{multirow}
\usepackage{graphicx}
\usepackage{bm}
\usepackage{amsthm}

\usepackage[normalem]{ulem}
\usepackage{tabularray}
\usepackage{cleveref}
\crefformat{section}{\S#2#1#3} 
\crefformat{subsection}{\S#2#1#3}
\crefformat{subsubsection}{\S#2#1#3}
\useunder{\uline}{\ul}{}

\definecolor{White}{rgb}{1, 1, 1}
\definecolor{Periwinkle}{rgb}{0, 0, 0}
\definecolor{myblue}{rgb}{0.82, 0.94, 0.75}
\definecolor{mygreen}{rgb}{0.64, 0.76, 0.68}
\definecolor{myyellow}{rgb}{0.88, 0.54, 0.35}
\definecolor{mygreen}{rgb}{0.68, 0.9, 0.8}
\definecolor{mypink}{rgb}{0.2, 0.87, 0.2}
\def\arrvline{\hfil\kern\arraycolsep\vline\kern-\arraycolsep\hfilneg}

\usepackage{enumitem}
\setlist[itemize]{align=parleft,left=0.5pt,itemsep=0.25pt,topsep=4pt}
\setlist[enumerate]{align=parleft,left=0.5pt,itemsep=0.25pt,topsep=4pt}

\usepackage{todonotes}

\title{Narrowing the Gap between Zero- and Few-shot \\ Machine Translation by Matching Styles}

\author{Weiting Tan, Haoran Xu, Lingfeng Shen, Shuyue Stella Li\\ \textbf{Kenton Murray, Philipp Koehn, Benjamin Van Durme, Yunmo Chen}\\[1em]
Johns Hopkins University\\
\texttt{\{wtan12,hxu64,lshen30,sli136,kenton,phi,vandurme,yunmo\}@jhu.edu}\\[1em]
}

\begin{document}
\maketitle
\begin{abstract}

Large language models trained primarily in a monolingual setting have demonstrated their ability to generalize to machine translation using zero- and few-shot examples with in-context learning. However, even though zero-shot translations are relatively good, there remains a discernible gap comparing their performance with the few-shot setting. In this paper, we investigate the factors contributing to this gap and find that this gap can largely be closed (for about 70\%) by matching the writing styles of the target corpus. Additionally, we explore potential approaches to enhance zero-shot baselines without the need for parallel demonstration examples, providing valuable insights into how these methods contribute to improving translation metrics.

\end{abstract}

%

\section{Introduction}


Recent advancements in large language models (LLMs) have revolutionized Natural Language Processing field as such models \citep[\textit{inter alia}]{openai2023gpt4, instrcutgpt, chowdhery2022palm, touvron2023llama} can easily adapt to a new task through prompt (in-context) learning where the task instruction and demonstrations (examples to guide LLMs on the task) are provided to the model. Such capability opens up new opportunities for machine translation, which is traditionally trained/fine-tuned on large amounts of parallel corpus \cite[\textit{inter alia}]{brown-etal-1993-mathematics, bahdanau2016neural, transformer, nllbteam2022language}. Recent work~\cite{vilar2022prompting, zhang2023prompting, jiao2023chatgpt, hendy2023good} has found that prompt-based methods perform well on language models trained primarily on monolingual data, rivaling state-of-the-art systems trained specifically for machine translation tasks on benchmark datasets.

\begin{figure}[t]
    \includegraphics[width=\linewidth]{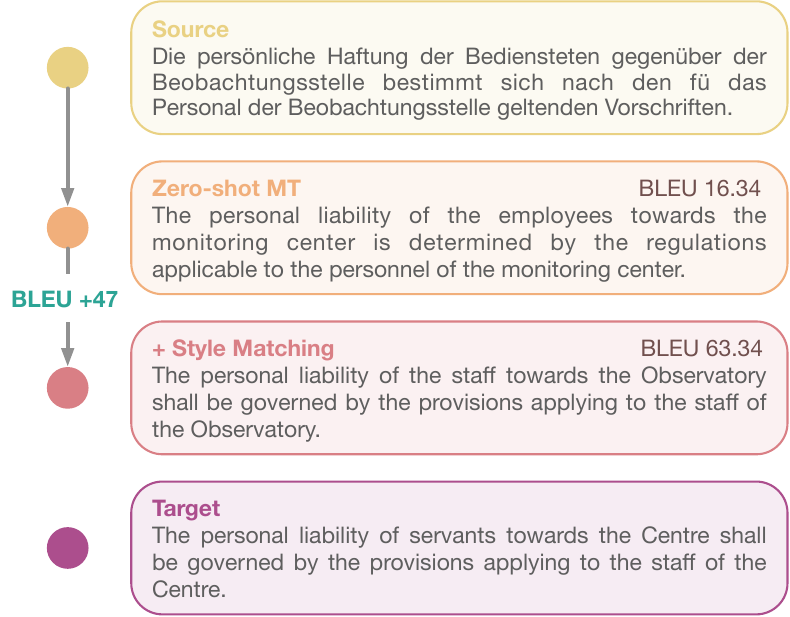}
    \caption{An example of translations under different settings. The zero-shot translation conveys the most semantic meaning of its source but lacks the ability to match with the style of its target sentence, resulting in a low BLEU score. With style-matching prompting, the LLM is able to generate a much better translation without accessing additional parallel examples.}
    \label{fig::teaser}
    \vspace{-5mm}
\end{figure}

Comparing zero-and few-shot outputs (\cref{sec::gap}), we observe a huge gap in terms of BLEU~\cite{papineni-etal-2002-bleu} score, despite the acceptable quality of zero-shot translations. As shown in \autoref{fig::teaser}, the zero-shot translation already conveys the meaning but the few-shot translation is better matched with the target (thus obtaining a better BLEU score). 

\begin{figure*}[t]
    \centering
    \includegraphics[width=1.0\linewidth]{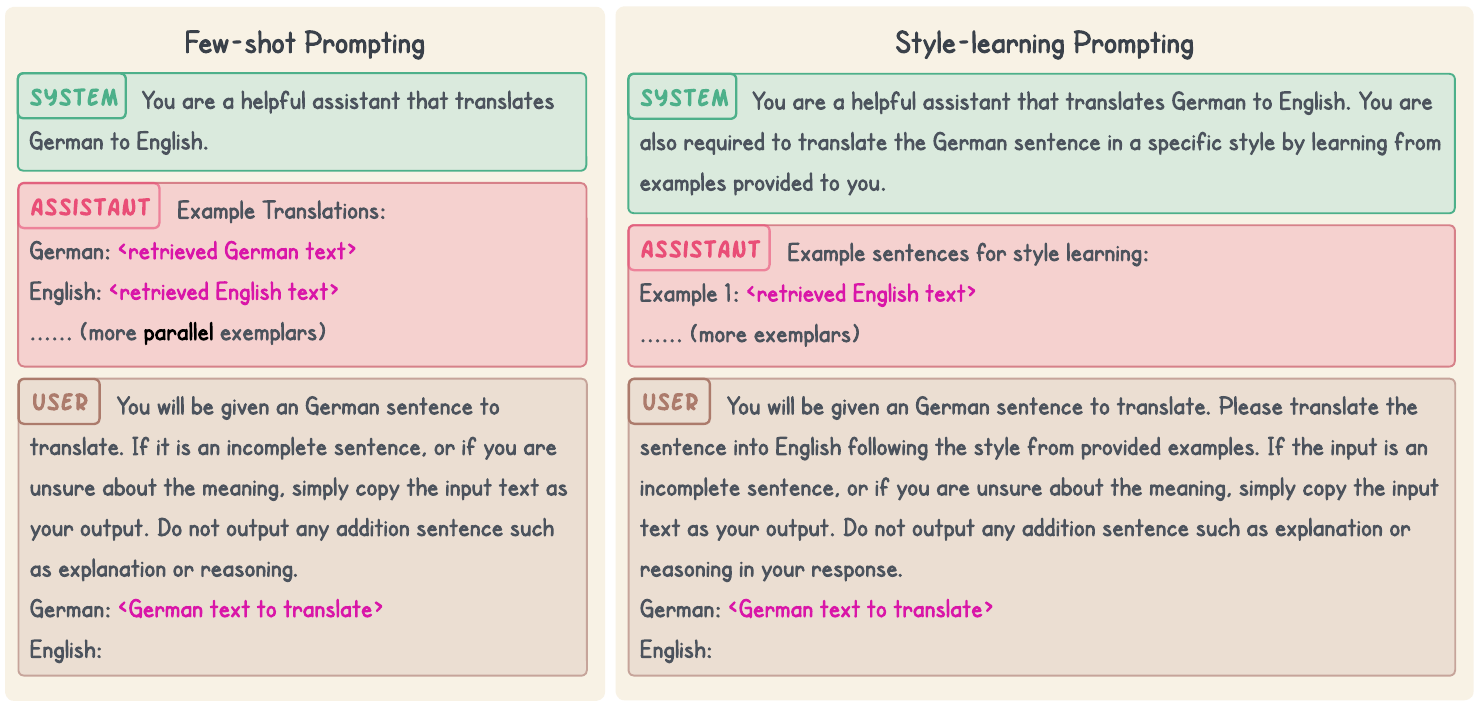}
    \vspace{-6mm}
    \caption{Prompt templates for instructing models to perform machine translation. \textit{Left} is the prompt template used for zero- and few-shot translation; in the few-shot scenario, source and target language pairs are required; in the zero-shot setting, example translations are not provided (\emph{i.e.} no assistant message is sent). \textit{Right} is the prompt template that incorporates style-learning instruction which only requires retrieving monolingual sentences from target corpora. In fact, style-learning prompting can also be considered as a kind of few-shot, but it only has access to target corpora. We denote ``few-shot'' as the experiment setup that requires parallel paired demonstrations.}
    \label{fig::fewshot_prompt}
    \vspace{-3mm}
\end{figure*}


We present qualitative and quantitative analysis on zero- and few-shot translation to understand their quality difference. Our evaluation (\cref{sec::gap}) reveals that the performance gap stems mostly from writing styles rather than semantics. This finding motivates us to quantify the degree of style match between translations and references (\cref{sec::analysis}) and develop a data-efficient style-learning prompting strategy (\cref{sec::style}).\footnote{~The style-learning prompting only has access to samples from target corpora. We denote ``few-shot'' as the experiment setup that requires paired parallel demonstrations.} Compared to few-shot translation, our prompting strategy relies solely on retrievals from in-domain target corpora and demonstrates effectiveness in closing approximately $70\%$ of the gap between zero-and few-shot translation. We provide an example in \autoref{fig::teaser} to illustrate the effectiveness of the proposed style-matching approach.

\section{The Gap between Zero- and Few-shot}\label{sec::gap}
For our experiments, we use the German-English data splits in \citet{aharoni-goldberg-2020-unsupervised}\footnote{~The dataset is originally collected by \citet{koehn-knowles-2017-six}.} that comprises IT, Subtitle, Law, Medical, and Koran domains. Due to the data quality issue, we keep Law, Medical, and Koran domains as they are relatively cleaner and representative of specialized domains other than generic daily language. For the details of setups and discussion about the dataset, we defer readers to \autoref{sec:setup}. 

\begin{figure*}[h]
    \includegraphics[width=1.0\linewidth, trim=2cm 2cm 2cm 2cm]{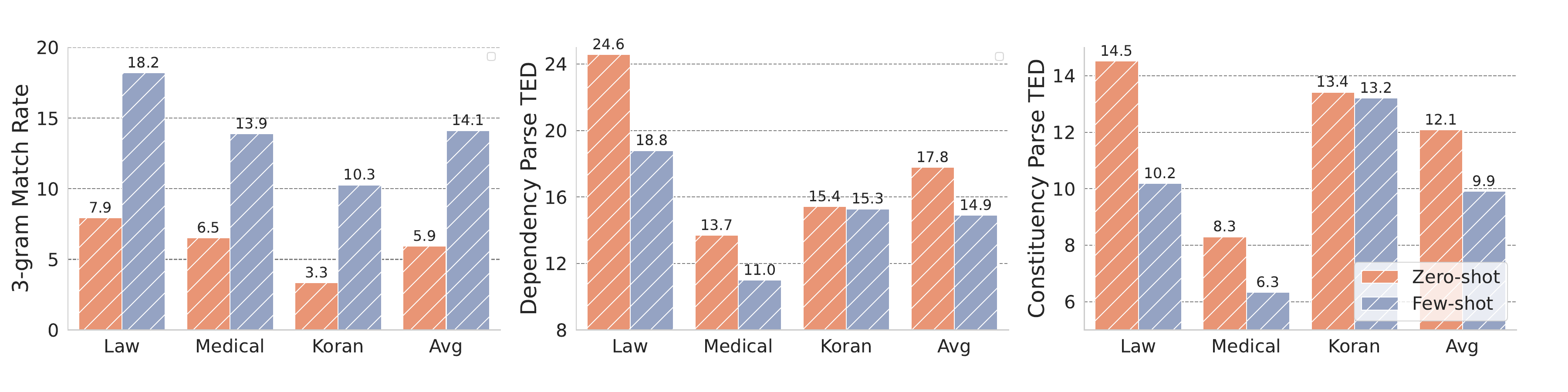}
    \vspace{-2mm}
    \caption{\textit{Left}: trigram overlap analysis between \emph{translations} and \emph{retrieved examples}; the gaps between zero- and few-shot indicate that choices of lexicons in few-shot have been greatly impacted by retrieved demonstrations. \textit{Mid}: averages of Tree Edit Distance (TED) between dependency parse trees from references and zero- or few-shot translations. Few-shot has a lower TED, showing that its grammar is more conformed with references. \textit{Right}: averages of TED between constituency parses from references and zero- or few-shot translations. Again, the few-shot results with lower TED are more conformed in syntactical structure with references.}
    \vspace{-2mm}
    \label{fig::n-gram}
\end{figure*}


Based on prior work on in-context learning \citep[\textit{inter alia}]{brown2020language, zhang2023prompting, vilar2022prompting, jiao2023chatgpt, hendy2023good}, we design prompts to instruct LLMs to perform zero- and few-shot translations. Specifically, for few-shot translation, we employ a retrieval-based prompting method \cite{hendy2023good, moslem2023adaptive, vilar2022prompting} that retrieves source and target pairs from the training corpus as in-context learning (ICL) demonstrations.


To facilitate this process, we create a prompt template illustrated in \autoref{fig::fewshot_prompt} \textit{Left}. In this template, the retrieved examples are embedded within the assistant message to aid in the few-shot translation. To retrieve relevant examples, we utilize the BM25 \cite{robertson1994okapi} retriever, with source sentences serving as queries. By searching the source language corpus, we identify the top matches and extract aligned parallel sentence pairs as demonstration examples.

\begin{table}[t]
\small
\centering
\begin{tabular}{ccc}
\toprule
\textbf{Method} & \textbf{BLEU} & \textbf{COMET}\\
\midrule
Vanilla Transformer & 41.4 & 80.4  \\
GPT3.5 0-shot & 31.4 & 82.1 \\
GPT3.5 5-shot & 47.6 & 84.6 \\
\bottomrule
\end{tabular}
\caption{We compare average BLEU scores for prompt-based methods (using gpt-3.5-turbo-0301 endpoint to translate from German to English) and vanilla transformer (a 6-layer encoder-decoder structure as introduced by~\citet{transformer}) trained on the in-domain dataset. We found that with few-shot translation, LLM could generate translation that achieves much better performance than 0-shot as well as domain-specific models.
}
\label{table::result_zeroshot_fewshot}
\vspace{-5mm}
\end{table}

We evaluate translation outputs using BLEU~\cite{papineni-etal-2002-bleu} and COMET~\cite{rei-etal-2020-comet}, and the results are presented in \autoref{table::result_zeroshot_fewshot}. 
In comparison to the few-shot performance (5-shot), zero-shot translation exhibits a noticeable gap of 13 points on the BLEU score. 
However, the difference is less significant when considering COMET, as the few-shot approach only outperforms the zero-shot counterpart by 2.5 points. To provide a better understanding of the COMET difference, we present a few examples in \autoref{fig::comet_diff} to illustrate how the score difference is reflected in translations.\footnote{~These examples serve as reference points to help readers qualitatively grasp the disparity in COMET scores.}

Examining the example with a COMET difference of 2.7 points, we observe that such a scale of difference, when mapped to lexicons, corresponds to only a few word changes, while the underlying semantics remain largely consistent with the reference. 
This minor difference in COMET scores indicates that zero-shot translations have already conveyed the semantic meaning of the source sentence, albeit with some variations in lexical choices and sentence structure. 

\section{How Much Does Style Contribute?}\label{sec::analysis}
We have observed that the lexical difference between zero-shot and few-shot translations is more noticeable than their semantic difference. Following the traditional linguistic definition of language style \citep[\textit{inter alia}]{mcdonald-pustejovsky-1985-computational, style_survey}, we can interpret this phenomenon as a distinction in writing style, where semantics remains consistent while aspects such as word choices, syntax, tones, etc., are altered.

To examine the potential impact of writing styles being learned through few-shot demonstrations, we investigate various aspects associated with writing styles to determine whether few-shot translation exhibits more consistent styles with target domains. We follow common techniques used in prior work \citep{andrea_garcia, Jankowska2017AuthorSA, Stamatatos2000AutomaticTC} to analyze style matching through three aspects: lexical overlaps, syntactical structures, and word dependencies. These aspects provide valuable insights into the output style of different systems. 



\subsection{Choice of Lexicons}\label{sec::ngram}
To understand how few-shot demonstrations affect lexicon usage in translations, we perform an analysis that identifies common n-gram presence between \emph{retrieved samples} and \emph{translations}. The match rate difference between zero- and few-shot indicates how much these examples change the choices of lexicons in translations.
We compute average trigram match rates between top-5 retrieved samples and zero- and few-shot translations (shown in \autoref{fig::n-gram} \textit{Left}).\footnote{~We explored a range of n-grams from unigram to 10-gram while observing a similar gap, thus presenting only the trigram results for simplicity.} Across all domains, the n-gram match rates are much higher for few-shot translations, indicating that few-shot translations are being steered to be lexical-wise more similar to demonstrations. 
Given that few-shot translations obtain better scores on BLEU in general, we attribute a non-trivial portion of performance gain to lexicons learned from its demonstrations, \emph{therefore better examples matched with target domains can potentially provide better guidance for few-shot translation}. A qualitative example is provided in \autoref{fig::ablation_example}.

\begin{figure*}
    \centering
    \includegraphics[width=1.0\linewidth, trim=0cm 0cm 0cm 2cm]{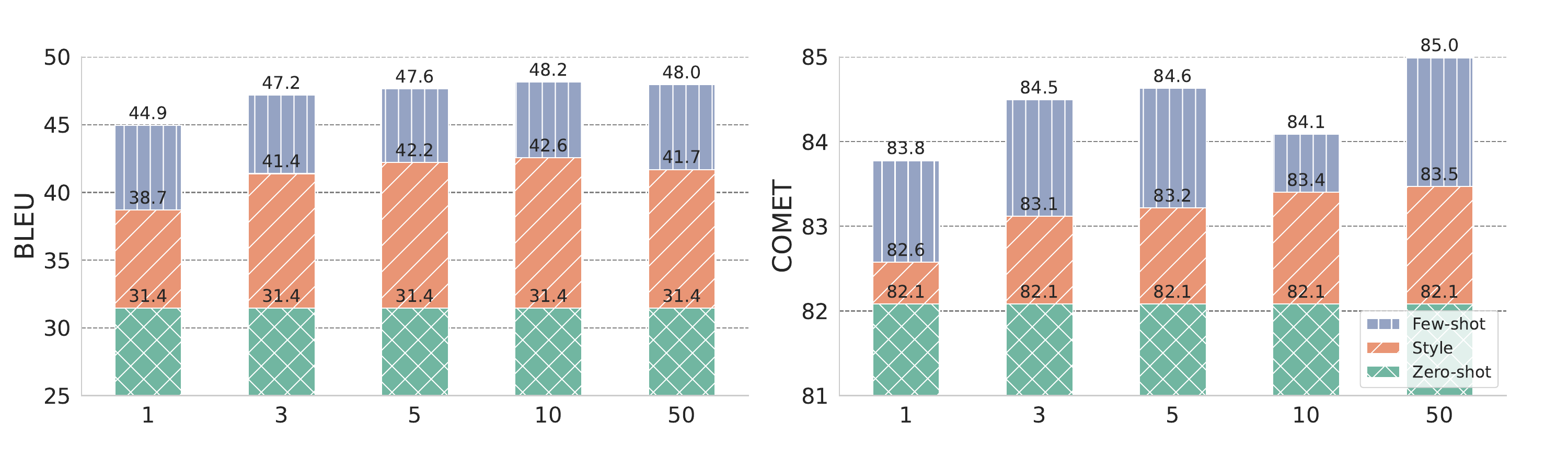}
    \vspace{-5mm}
    \caption{Results of few-shot, zero-shot, and zero-shot with style-learning prompting evaluated using BLEU and COMET. The x-axis is the number of demonstrations used in the prompt; in few-shot, they are parallel sentence pairs, whereas in zero-shot with style learning, they are just target samples. The y-axis is the metric score. Both few-shot and style-learning prompting improve from the zero-shot baseline substantially even with only one example.}
    \label{fig::result}
    \vspace{-4.5mm}
\end{figure*}

\subsection{Organization of Sentence Structures}\label{sec::ted}

\paragraph{Word Semantic Relations} The degree of style matching between translations and the references can be measured by comparing the roles of different words under dependency parse trees (shown in \autoref{fig::n-gram} \textit{Mid}).\footnote{~Dependency parsing and constituency parsing were performed using the spaCy toolkit (\url{https://spacy.io/})} 
We quantify the difference between parses using Tree Edit-Distance (TED), where smaller mean closer (or more similar) between trees.\footnote{~TED is computed using the Zhang-shasa algorithm \cite{zss}, a classical algorithm for computing similarity between two parses.}
Few-shot translations exhibit a smaller overall TED, indicating closer alignments in word semantic relations with references.

\paragraph{Syntactical Structure Similarity} We also apply TED on constituent trees to understand similarities of syntactical structures between translations and references.\footnote{~We removed nodes at the bottom level so that TED would not reflect lexical differences.} Our results show that few-shot translations also demonstrates higher similarities in syntactical structures with target domains.


\section{Narrowing the Gap via Matching Styles}\label{sec::style}

From \cref{sec::analysis}, we demonstrated that few-shot conforms to the target style better than zero-shot translation. Motivated by this observation, we investigate approaches to mitigate the style difference so that zero-shot translation can be improved without access to additional parallel examples. 

We propose a novel style-learning prompting approach that leverages target corpora, as illustrated in \autoref{fig::fewshot_prompt} \textit{Right}. Our hypothesis is that style-relevant information primarily resides in the in-domain target corpus (in our case, English). Therefore, by retrieving samples from the corresponding monolingual corpus, the style-learning instruction would enhance style consistency in translation. This idea is in the same spirit as prior work in test time domain adaptation \citep[\textit{inter alia}]{singh2022addressing,zheng2022nonparametric,zhang-etal-2022-iterative,hu2019domain}, where target domain information is used to guide model inference. In our case, we allow language models (LLMs) to access examples from the target corpora, enabling them to adjust their language styles.

Our approach has two steps: first, we employ zero-shot translation; then we take the translation output as a query to retrieve examples from the target language's corpus and \emph{add them to the original translation prompt with a style-learning instruction}. By revising prompts with target examples, the approach allows us to align styles in translations without source-side demonstrations.

The performance comparison\footnote{~The breakdowns for domains are provided in \autoref{fig::result-full}.} among few-shot, zero-shot, and zero-shot with style matching is shown in \autoref{fig::result}, where the number of demonstrations ranges from 1 to 50. From the results, few-shot prompting exhibits the best performance by leveraging both source and target information. Our style-learning prompts largely improve the zero-shot performance, reducing the gap by approximately 70\%. To further validate these findings, we conducted experiments using the Llama2-7b model~\cite{touvron2023llama}, fixing the number of demonstrations at 5 for both German-English and English-German translation directions. As depicted in \autoref{table::llama_result}, the trend remains consistent, demonstrating that style-learning prompts effectively bridge the gap between zero-shot and few-shot translations. Interestingly, Llama2-7b shows less improvement with style-learning prompts and has a notably lower zero-shot baseline compared to GPT3.5, suggesting that the impact of style-learning varies among models, dependent on their initial translation capabilities.


\begin{table*}[h]
\centering
\small
\setlength{\tabcolsep}{6pt} 
\renewcommand{\arraystretch}{1.2} 
\begin{tabular}{lccccccccccc}
    \toprule
    \multirow{2}{*}{Direction} & \multirow{2}{*}{Method} & \multicolumn{2}{c}{Law} & \multicolumn{2}{c}{Med} & \multicolumn{2}{c}{Koran} & \multicolumn{2}{c}{Average} \\
    \cmidrule(lr){3-4} \cmidrule(lr){5-6} \cmidrule(lr){7-8} \cmidrule(lr){9-10}
     & & GPT3.5 & Llama2 & GPT3.5 & Llama2 & GPT3.5 & Llama2 & GPT3.5 & Llama2 \\
    \midrule
    \multirow{3}{*}{DE $\rightarrow$ EN} & Zero-shot & 37.2 & 32.5 & 44.5 & 39.2 & 17.9 & 14.4 & 33.2 & 28.7 \\
    & Style & 48.9 & 46.1 & 57.8 & 44.6 & 19.8 & 15.5 & 42.2(\textcolor{red}{+9.0}) & 35.4(\textcolor{red}{+6.7}) \\
    & Few-shot & 59.3 & 58.9 & 64.1 & 58.9 & 20.2 & 16.6 & 47.9(\textcolor{red}{+14.7}) & 44.8(\textcolor{red}{+16.1}) \\
    \midrule
    \multirow{3}{*}{EN $\rightarrow$ DE} & Zero-shot & 32.1 & 22.6 & 38.8 & 28.7 & 12.8 & 8.3 & 27.9 & 19.9 \\
    & Style & 41.1 & 28.0 & 47.8 & 32.9 & 23.5 & 14.9 & 37.4(\textcolor{red}{+9.5}) & 25.2(\textcolor{red}{+5.3}) \\
    & Few-shot & 49.6 & 48.2 & 54.5 & 51.3 & 20.3& 24.8 & 41.4(\textcolor{red}{+13.5}) & 41.5(\textcolor{red}{+21.6}) \\
    \bottomrule
    \end{tabular}
\caption{GPT3.5 and Llama2-7b's performance across different domains with zero-shot, few-shot, and style learning prompting. The style-learning prompting strategy improves both GPT3.5 and Llama2's performance with only monolingual information.}
\label{table::llama_result}
\end{table*}

\begin{table*}[h]
\small
\centering
\begin{tabular}{ccccccccccc}
\toprule
\multirow{2}{*}{Prompt Style} & \multirow{2}{*}{} & \multicolumn{4}{c}{Fewshot} & \multirow{2}{*}{} & \multicolumn{4}{c}{Style} \\
\cmidrule{3-6} \cmidrule{8-11}
& & 1-shot & 5-shot & 10-shot & Avg & & 1-shot & 5-shot & 10-shot & Avg \\
\midrule
Tier 0 & & 43.4 & 47.9 & 48.1 & 44.9 & & 38.5 & 42.2 & 42.8 & 41.2 \\
Tier 1 & & 36.7 & 38.8 & 39.7 & 38.4 (-14\%) & & 34.6 & 36.1 & 36.6 & 35.7 (-13\%) \\
Tier 2 & & 36.2 & 38.1 & 39.2 & 37.8 (-16\%) & & 34.4 & 35.8 & 36.4 & 35.5 (-14\%) \\
Tier 3 & & 36.2 & 38.1 & 39.2 & 37.8 (-16\%) & & 34.2 & 35.7 & 36.0 & 35.3 (-14\%) \\
\bottomrule
\end{tabular}
\caption{Prompt performance (averaged across domains; more details in \autoref{table:ablation-full}) with retrieved samples of varying quality. Using a single demonstration from Tier 0 is more effective than using 10 demonstrations from Tier 1.}
\vspace{-3mm}
\label{table:ablation-tier}
\end{table*}

Overall, few-shot prompting is still more effective and we attribute such advantages of few-shot over style-learning prompting to: 
\begin{itemize}
    \item The few-shot setting has access to additional paired demonstrations, which can provide better guidance, such as language alignment, compared to monolingual examples employed in style-learning prompting.
    \item The retriever queries the target corpus with the potentially noisy translation generated from zero-shot translation for style-learning prompting.
\end{itemize}

Despite the remaining performance gap, our style-learning prompting method demonstrates increased data efficiency and adaptability of the language model to new domains where only monolingual resources are available. 


\paragraph{Ablation on Retrieval Quality}
\begin{figure}[h]
    \centering
    \includegraphics[width=0.96\linewidth]{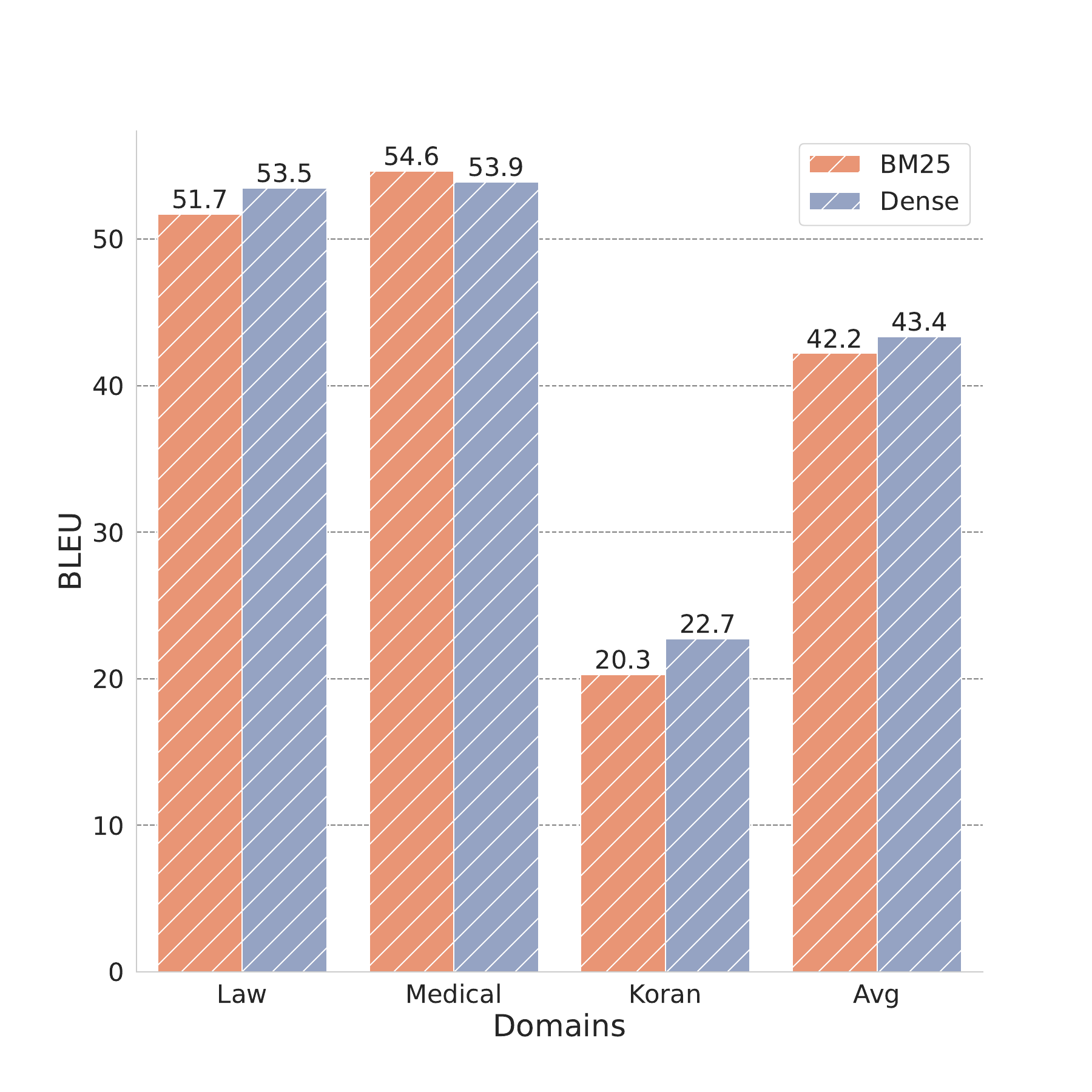}
    \vspace{-3mm}
    \caption{BM25 vs Dense retriever's effect on style-learning prompting's translation quality (with k=5 examples)}
    \label{fig::bm_dense}
    \vspace{-3mm}
\end{figure}
For both few-shot and our style-learning prompting, the quality of retrieved examples is important. We retrieved the 100 most similar samples with BM25 and separated them into four groups (tier 0 to 3, with 0 being the most similar group). As shown in \autoref{table:ablation-tier}, for both few-shot and style-learning, they benefit more from just a single example from tier 0 than from the 10 samplers from tier 1, indicating the importance of retrieval quality. For the full results, please refer to \autoref{table:ablation-full}. 
In \autoref{fig::bm_dense}, we also show that a dense retriever (based on sentence-transformer) obtains better demonstrations for the style-learning prompt.
Though improving retrievers is orthogonal to our contribution of prompting strategy in this paper, we have shown that retriever is indeed essential in our pipeline for demonstration curation. We envision that a retriever that specifically attends to fine-grained style might be developed to further improve the style-learning prompting performance.


\section{Conclusion}
In this paper, we investigate the prompt-based zero-and few-shot translation performance on different domains. We attribute the performance gap to the style difference, and through our analysis, we verified that few-shot translation benefits from its retrieved samples by learning the target domain's style. We then design a style-learning prompting strategy that only requires monolingual information and could achieve about $70\%$ performance gain (for GPT3.5) compared to its few-shot counterpart.

\section*{Limitations}
\paragraph{Model:} This paper's experiment is currently based on the GPT3.5-0301 snapshot and LLaMa2-7b model, and we will include ICL performance from other LLMs in the future. However, we anticipate that the observed trends and findings would likely extend to other LLMs.  
\paragraph{Translation Dataset:} The translation datasets used in this study exclusively involve German-to-English translation, which represents a high-resource language pair. Consequently, the results may not be directly applicable to low-resource translation tasks or languages in which LLMs do not perform well. We leave the investigation of low-resource languages to future work.




\bibliography{anthology,custom}
\bibliographystyle{acl_natbib}

\newpage

\onecolumn

\appendix

\section{Experiment Setup}
\label{sec:setup}
\paragraph{Model:} For all our experiments, we use the publicly available ChatGPT 
\textit{gpt-3.5-turbo-0301}\footnote{~~\url{https://platform.openai.com/docs/guides/chat}} model. For hyper-parameters, we use \textit{temperature} 0.3 throughout our experiments, and we set all other hyper-parameters using their default values.

\paragraph{Evaluation Metrics:} We employ BLEU \cite{papineni-etal-2002-bleu} and COMET \cite{rei-etal-2020-comet} to compare the reference and hypothesis. 

\paragraph{Datasets} For the dataset, we rely on the domain data splits in \citet{aharoni-goldberg-2020-unsupervised}, originally collected by \citet{koehn-knowles-2017-six}. This corpus includes five domain-specific (\texttt{Law}, \texttt{Medical}, \texttt{Koran}, \texttt{IT}, and \texttt{Subtitles}) datasets for German-English translation. We did not use \texttt{IT} and \texttt{Subtitles} domains because the sentences in these two domains are noisy and short (with only 9 words/sentence for \texttt{IT} and 8.2 words/sentence for \textit{Subtitles}). From our preliminary study, we also found that lots of noisy sentences are not filtered out in this dataset. Therefore we used ChatGPT to automatically filter the noisy pairs from the test set (we are not exposed to the dataset and for the details, please refer to \cref{sec::filter}). In this paper, we report evaluation results based on the cleaned dataset\footnote{~~The cleaned dataset will be released at: anonymized.com}.
\begin{table}[h]\small
    \centering
    \begin{tabular}{ccc}
    \toprule
     Domain & \#Sentence & \#Words/Sentence  \\
     \midrule
     Law & 1907 & 28.6 \\
     Medical & 1665 & 16.9 \\
     Koran & 1629 & 20.3 \\ 
     \bottomrule
    \end{tabular}
    \caption{Test data statistics after applying GPT3.5 filter following \cref{sec::filter}}
    \label{tab:my_label}
\end{table}

\paragraph{Data Cleaning with GPT-3.5}\label{sec::filter}
To clean the test set without being exposed to the sample, we first perform zero-shot translation on the devset following \autoref{fig::fewshot_prompt}. Then we rank the generated outputs by their BLEU score and take the 20 worst German-English samples (most of which are low-quality because of noise in the sentence pair, as you can find in \autoref{text:filter-medical}). Note that we take the original source-target sentence pairs instead of the zero-shot output because we only use zero-shot's performance as a scorer to obtain the worst samples. Then we prompt GPT-3.5 to evaluate the quality of these ill-formed pairs and get the evaluation results. The criteria section of the prompt shown in \autoref{text:filter-law}, \autoref{text:filter-koran}, \autoref{text:filter-medical} comes from GPT-3.5's evaluation results. Lastly, we use these prompts to ask GPT-3.5 whether each test sentence pair is a good translation and we filter out those pairs that GPT-3.5 predicts "No".



\begin{table*}[htbp]
\small
\centering
\begin{tabular}{cccccccccccc}
\toprule
Prompt Style: & & \multicolumn{4}{c}{Fewshot} & & \multicolumn{3}{c}{Style-Transfer} \\
\toprule
Domain & Tier & 1-shot & 5-shot & 10-shot & Avg & & 1-shot & 5-shot & 10-shot & Avg \\
\midrule
& Tier 0 & 43.4 & 47.9 & 48.1 & 44.9 &
& 38.5 & 42.2 & 42.8 & 41.2\\
& Tier 1 & 36.7 & 38.8 & 39.7 & 38.4 (-14\%) & 
& 34.6 & 36.1 & 36.6  & 35.7 (-13\%) \\
Average & Tier 2 & 36.2 & 38.1 & 39.2  & 37.8 (-16\%) &
& 34.4 & 35.8 & 36.4 & 35.5 (-14\%)\\
& Tier 3 & 36.2 & 38.1 & 39.2 & 37.8 (-16\%)& 
& 34.2 & 35.7 & 36.0  & 35.3 (-14\%) \\
\midrule
& Tier 0 & 54.9 & 59.3 & 60.0 & 58.1 &
& 45.5 & 48.9 & 48.9 & 47.8\\
& Tier 1 & 43.8 & 47.1 & 48.2 & 46.4 (-20\%) & 
& 40.7 & 42.3 & 42.8  & 41.9 (-12\%) \\
Law & Tier 2 & 42.7 & 45.4 & 47.4  & 45.1 (-22\%) &
& 40.3 & 42.1 & 42.5 & 41.7 (-13\%)\\
& Tier 3 & 42.8 & 46.5 & 47.6 & 45.6 (-21\%)& 
& 40.0 & 42.0 & 42.8  & 41.6 (-13\%) \\
\midrule
& Tier 0 & 57.7 & 64.1 & 63.6 & 57.5 &
& 50.2 & 57.8 & 59.4 & 55.8 \\
& Tier 1 & 46.4 & 48.5 & 49.3 & 48.1 (-16\%) &
& 44.3 & 45.9 & 47.0 & 45.7 (-18\%) \\
Medical & Tier 2 & 46.5 & 48.0 & 48.7 & 47.7 (-17\%) & 
& 44.3 & 45.7 & 46.4 & 45.5 (-18\%) \\
& Tier 3 & 46.1 & 47.5 & 48.4 & 47.3 (-17\%) &
& 43.9 & 45.7 & 45.4 & 49.0 (-19\%) \\
\midrule
& Tier 0 & 17.6 & 20.2 & 20.5 & 19.0 &
& 19.7 & 19.8 & 20.2 & 19.9 \\
& Tier 1 & 19.8 & 20.6 & 21.7 & 20.7 (+9\%)&
& 18.7 & 20.0 & 20.1 & 19.6 (-1\%) \\
Koran & Tier 2 & 19.2 & 20.9 & 21.7 & 20.6 (+8\%) &
& 18.5 & 19.6 & 20.2 & 19.4 (-2\%) \\
& Tier 3 & 19.4 & 20.6 & 21.1 & 20.4 (+7 \%) &
& 18.5 & 19.5 & 19.9 & 19.3 (-3\%) \\
\bottomrule
\end{tabular}
\caption{Prompt performance with retrieved samples of varying quality}
\label{table:ablation-full}
\end{table*}

\section{Ablation on ICL Demonstrations' Quality}\label{sec::ablation}
To further understand the impact of demonstrations' quality, we ablate on the quality of samples retrieved with the lexical retriever -- BM25. We use BM25 retriever to obtain 100 most similar samples from the training corpus for few-shot translation. From the most similar 100 samples, we craft four different chunks, each of 25 samples. We rank these chunks of retrieved data as tier 0, tier 1, tier 2, and tier 3 where tier 0 consists of 25 samples that have the highest BM25 scores, tier 1 has the next 25 samples with the highest similarity score, and so on. From \autoref{table:ablation-full}, we see that the performance of prompt (averaged across domain) degrades by $14\%$ from tier 0 to tier 1, showing that the quality of high-similarity samples that are essential for the high performance. Our finding provides a different view compared to prior work \cite{hendy2023good, vilar2022prompting} which finds random and searched samples (with embedding-based k-nearest-neighbor~\cite{cover1967nearest} methods) have no effect on the performance. We hypothesize that such difference comes from the nature of the task and dataset since our domain-specific dataset has more conformed styles while prior works tested on general benchmark datasets from WMT, which makes style-matching harder with few examples.



\begin{figure*}
    \includegraphics[width=1.0\linewidth]{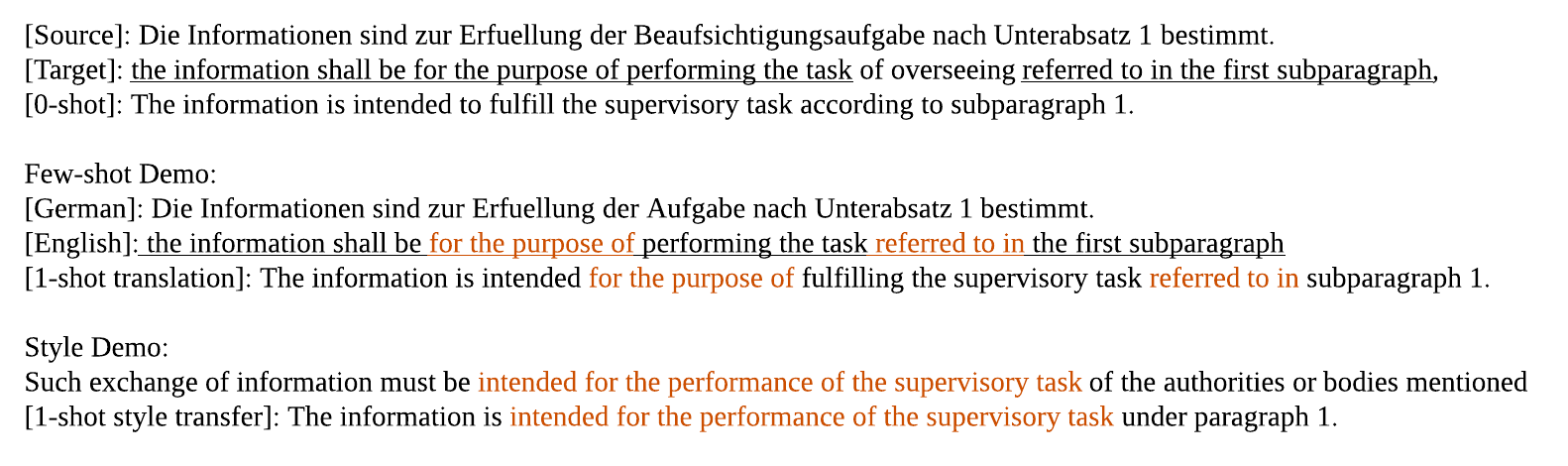}
    \caption{The example shown here is an illustration of the improvement from few-shot and style-learning translation. The text in orange is the text that GPT3.5 learns from the demonstration, which results in better lexical overlap and higher syntactic similarity with the reference text. In this example, the Few-shot prompt's demo has better quality as it overlaps more with the target sentence (in fact, the whole sentence can be found in the target sentence, as shown with the underline). The retriever for the few-shot prompts has better quality as it is able to utilize both the source and target language's corpus.}
    \label{fig::ablation_example}
\end{figure*}

\begin{figure*}
    \centering
    \includegraphics[width=1.0\linewidth]{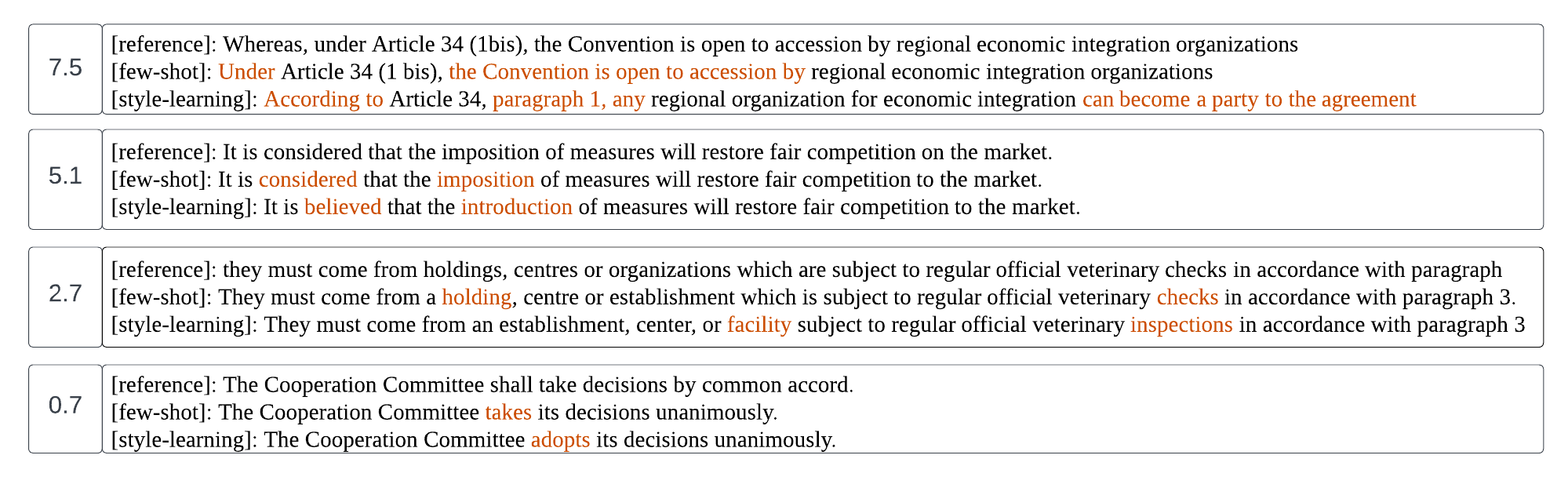}
    \caption{Here we provide 4 examples of comparison of COMET scores between few-shot and style-learning prompting outputs. The number on the left is the COMET score difference and the highlighted words are the major difference between the two outputs. We see that all these samples have similar semantics and the difference is mostly from lexical choice (except for the 7.5 cases where sentence structure is also changed but the semantics is still very similar). Therefore, we believe zero-and few-shot outputs' performance gap originates from styles rather than semantics.}
    \label{fig::comet_diff}
\end{figure*}

\begin{figure*}
    \centering
    \includegraphics[width=1.0\linewidth]{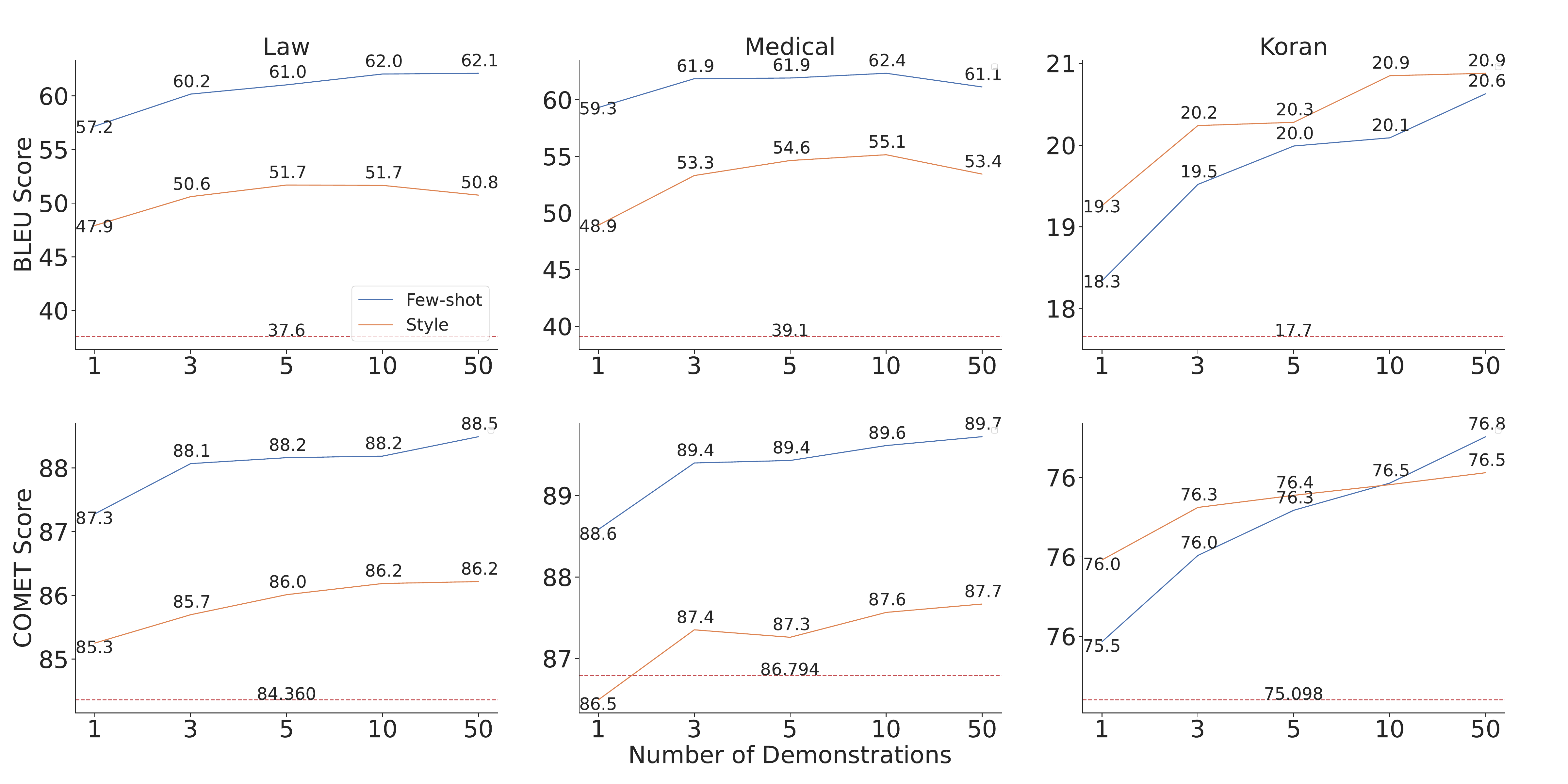}
    \caption{Full Results of few-shot and style-learning prompting across three domains, evaluated by BLEU and COMET (COMET score is multiplied by 100 for better visualization). The x-axis is the number of demonstrations used in the prompt and the y-axis is the evaluation result. The dashed red line is the performance of zero-shot translation and we see that both prompts improve the domain-specific translation even just given 1 demo.}
    \label{fig::result-full}
\end{figure*}

\begin{figure*}[h]
    \includegraphics[width=1.0\linewidth, trim=2cm 2cm 2cm 2cm]{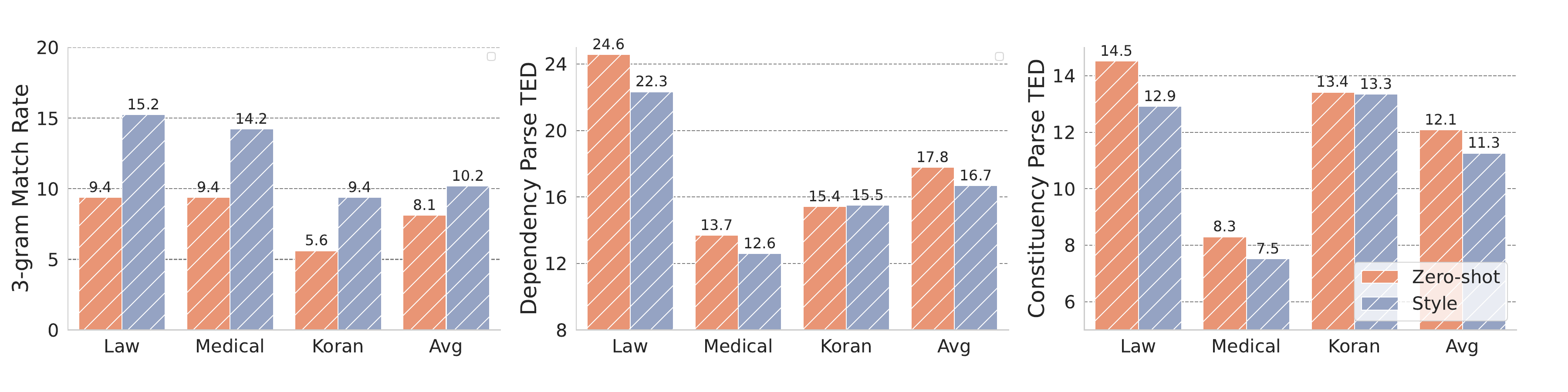}
    \caption{\textit{Left}: 3-gram overlap analysis between \emph{translations} and \emph{retrieved examples}. Compared to zero-shot, the output from style-learning prompting has a much higher match rate with retrieved samples. \textit{Mid}: Average Tree Edit Distance (TED) between the dependency parse tree of reference and outputs from zero-shot or style-learning prompting. Style-learning prompting results in a lower TED, showing its more grammar with the reference. \textit{Right}: Average TED between the constituency parse of reference and zero-shot/style-learning outputs. Again, the style-learning prompting achieves lower TED and has a more conformed syntactical structure with the reference.}
    \vspace{-5mm}
    \label{fig::n-gram-style}
\end{figure*}

\begin{table*}[ht]
    \centering
    \small
    \noindent\fbox{%
    \begin{minipage}{\dimexpr\linewidth-3\fboxsep-2\fboxrule} 
You are a human evaluator who judges the quality of parallel data for German-English translation. Below are some examples of low-quality parallel data along with the criteria for filtering them out.
[German]: DIESE HINTERBLIEBENENVERSORGUNG ENTSPRICHT :

[English]: THAT PENSION SHALL BE EQUAL TO A PERCENTAGE OF THE PENSION ACCRUING TO THE MEMBER OR FORMER MEMBER OF THE COMMISSION OR OF THE COURT UNDER ARTICLE 9 AT THE DATE OF DEATH, NAMELY:

[Criteria]: The translation is overly verbose and does not accurately convey the meaning of the German sentence. The English translation includes unnecessary repetitions and lacks clarity.\\

[German]: 3 2 0 0 BAR Information und Veröffentlichung BAR 200000 BAR BAR 200000 BAR

[English]: 3 2 0 0 BAR Information and publishing BAR 200000 BAR BAR 200000 BAR

[Criteria]: The translation includes non-translated elements (such as "BAR") that do not provide any meaningful information in the target language. It appears to be a result of incorrect processing or formatting.\\

[German]: - gelegentlich erfolgen,

[English]: - are of an occasional nature,

[Criteria]: The translation fails to capture the meaning of the German sentence accurately. It provides a more general interpretation, which does not convey the intended sense of occasional occurrences.\\

[German]: a) höher sein als die Einheit,

[English]: (a) be greater than 1;

[Criteria]: The translation does not accurately convey the meaning of the German sentence. It provides a different interpretation, suggesting that the value should be greater than 1, whereas the German sentence simply states "higher than the unit."\\

[German]: - Exportação para a Polónia.

[English]: - Exportação para a Polónia.

[Criteria]: The translation is not in English but rather includes Portuguese words. It appears to be a mistake or a mix-up between different language pairs."""\\

Given the criteria above, is following sentence pair a good translation? Output Yes if it is a good translation, output No if it is a bad translation.\\

[Germain]: <test source sentence>
[English]: <test target sentence>
    \end{minipage}
}
    \caption{Filtering Prompt for Law Domain}
    \label{text:filter-law}
\end{table*}
\begin{table*}[ht]
    \centering
    \small
    \noindent\fbox{%
    \begin{minipage}{\dimexpr\linewidth-3\fboxsep-2\fboxrule} 

You are a human evaluator who judges the quality of parallel data for German-English translation. Below are some examples of low-quality parallel data along with the criteria for filtering them out.\\

[German]: Brivudin behandelt wurden.

[English]: Xeloda should also not be used in the following groups...

[Criteria]: Incompleteness or Omission: When important information is missing in the translation or not adequately conveyed.\\

[German]: 12/23

[English]: 12/ 22

[Criteria]: Irrelevant or Unrelated Information: When the translation includes information that is not relevant to the original sentence.

[German]: Midazolam zur Injektion).

[English]: Examples include the cholesterol-reducing agent atorvastatin, the antibiotics...

[Criteria]: Sentence Structure and Syntax: When the translation has incorrect word order or sentence structure.\\

[German]: Hyperthyreose

[English]: Thyroid dysfunction most often presenting as hypothyroidism or hyperthyroidism

[Criteria]: Incorrect Terminology or Inaccurate Vocabulary: When the translation uses incorrect or inaccurate terms.\\

[German]: 1120

[English]: 112 Further training, language courses...

[Criteria]: Mistranslation or Misinterpretation: When the translation conveys a different meaning than the original sentence.\\

[German]: Die Wirksamkeit von Revasc als Gerinnungshemmer wurde in vier Studien...

[English]: ho death/ re-MI at Day 30 were not statistically different...

[Criteria]: Inconsistent Terminology: When the translation uses inconsistent or contradictory terms.\\

[German]: 7,9 µmol/l.

[English]: Ki values in human liver microsomes were 27, 7.5 and 7.9 µmol/ l, respectively.

[Criteria]: Grammatical Errors: When the translation contains grammatical mistakes or incorrect usage of language rules.\\

[German]: Uber diesen Minimalwert hinausgehende

[English]: The recommendation of a minimum yield of 2.0 x 106 CD34+ cells/ kg

[Criteria]: Lack of Clarity or Ambiguity: When the translation is unclear or ambiguous, making it difficult to understand the intended meaning.\\

Given the criteria above, is following sentence pair a good translation? Output Yes if it is a good translation, output No if it is a bad translation.\\

[Germain]: <test source sentence>
[English]: <test target sentence>

    \end{minipage}
}
    \caption{Filtering Prompt for Koran Domain}
    \label{text:filter-koran}
\end{table*}
\begin{table*}[ht]
    \centering
    \small
    \noindent\fbox{%
    \begin{minipage}{\dimexpr\linewidth-3\fboxsep-2\fboxrule} 
You are a human evaluator who judges the quality of parallel data for German-English translation. Below are some examples of low-quality parallel data along with the criteria for filtering them out.\\

[German]: Doch, mit Sicherheit!

[English]: Why not?

[Criteria]: The English translation does not capture the meaning of the German sentence, which should be translated as "Certainly, for sure!" or "Yes, definitely!" instead of "Why not?"\\

[German]: Dahin werdet ihr kommen müssen.

[English]: The Prophets like Eisa and Uzair who were worshipped are exempt from this, and so are Maryam, and trees and the moon etc.)

[Criteria]: The English translation is completely unrelated to the original German sentence. It introduces a different topic and provides irrelevant information. A more accurate translation would be "There you will have to go."\\

[German]: " Wir sind zugrunde gerichtet!

[English]: (And say:) "We have fallen into debt;

[Criteria]: The English translation does not convey the intended meaning of the German sentence. A more coherent translation would be "We are destroyed!" or "We are ruined!"\\

Given the criteria above, is following sentence pair a good translation? Output Yes if it is a good translation, output No if it is a bad translation.\\

[Germain]: <test source sentence>
[English]: <test target sentence>

    \end{minipage}
}
    \caption{Filtering Prompt for Medical Domain}
    \label{text:filter-medical}
\end{table*}

\end{document}